\documentclass[journal]{IEEEtran}
\usepackage{amsmath, amsfonts, amsthm}

\ifCLASSINFOpdf
\else
   \usepackage[dvips]{graphicx}
\fi
\usepackage{url}
\usepackage{graphicx}

\usepackage[style=ieee,natbib=true,backend=bibtex,doi=false,isbn=false,url=false,eprint=false,mincitenames=1,maxcitenames=1,uniquelist=false,alldates=year, maxbibnames=10]{biblatex}
\AtEveryBibitem{\clearlist{language}\clearname{editor}\clearfield{note}}
\bibliography{refs_updated} 

\usepackage{xcolor, soul}
\sethlcolor{pink}
\usepackage{array}
\usepackage{caption}
\usepackage{subcaption}
\usepackage{booktabs}
\usepackage{float}

\usepackage{amsmath,amsfonts,bm}

\def\1{\bm{1}}
\newcommand{\train}{\mathcal{D}}

\def\vb{{\bm{b}}}

\def\vv{{\bm{v}}}

\def\vx{{\bm{x}}}
\def\vy{{\bm{y}}}
\def\vz{{\bm{z}}}

\def\mI{{\bm{I}}}

\def\mK{{\bm{K}}}

\def\mP{{\bm{P}}}

\def\mU{{\bm{U}}}
\def\mV{{\bm{V}}}
\def\mW{{\bm{W}}}
\def\mX{{\bm{X}}}

\def\mSigma{{\bm{\Sigma}}}

\DeclareMathAlphabet{\mathsfit}{\encodingdefault}{\sfdefault}{m}{sl}
\SetMathAlphabet{\mathsfit}{bold}{\encodingdefault}{\sfdefault}{bx}{n}

\def\gF{{\mathcal{F}}}

\def\gN{{\mathcal{N}}}

\def\gU{{\mathcal{U}}}

\def\gX{{\mathcal{X}}}

\def\sZ{{\mathbb{Z}}}

\newcommand{\R}{\mathbb{R}}

\def\vbeta{{\bm{\beta}}}
\def\valpha{{\bm{\alpha}}}

\newcommand{\dimFeatures}{p}
\newcommand{\dimLatent}{d}
\newcommand{\dimRFF}{r}
\newcommand{\numSamples}{n}

\definecolor{red}{RGB}{215,25,28}
\definecolor{orange}{RGB}{253,174,97}
\definecolor{yellow}{RGB}{255,255,191}
\definecolor{lightblue}{RGB}{171,217,233}
\definecolor{darkblue}{RGB}{44,123,182}
\definecolor{lightgreen}{RGB}{178,223,138}
\definecolor{darkgreen}{RGB}{51,160,44}

\newcommand{\bigO}{\mathcal{O}}

\usepackage{hyperref}       
\hypersetup{
    colorlinks=true,
    citecolor=blue,
    linkcolor=red,
    filecolor=magenta,      
    urlcolor=violet}

\newcommand{\beginsupplement}{%
        \setcounter{table}{0}
        \renewcommand{\thetable}{A-\arabic{table}}%
        \setcounter{figure}{0}
        \renewcommand{\thefigure}{A-\arabic{figure}}%
     }

\usepackage{tikz}
\usepackage{pgfplots}
\pgfplotsset{compat=1.14}
\usepgfplotslibrary[fillbetween] 
\usepgfplotslibrary{external}
\usetikzlibrary{arrows.meta,calc,chains,shapes.geometric}

\newif\ifarxiv
\arxivtrue 

\begin{document}

\title{Invertible Kernel PCA with\\ Random Fourier Features}

\author{
	Daniel Gedon, Ant\^onio H. Ribeiro, Niklas Wahlström, and Thomas B. Schön, \IEEEmembership{Senior Member, IEEE}
    \thanks{Manuscript submitted 27 February 2023.
    This work was partially supported by the Wallenberg AI, Autonomous Systems and Software Program (WASP) funded by the Knut and Alice Wallenberg Foundation; by Kjell och M{\"a}rta Beijer Foundation; and by the Swedish Research Council (VR) via the project Physics-informed machine learning (registration number: 2021-04321)}
    \thanks{All authors are with the Department of Information Technology, Uppsala University, 751~05~Uppsala, Sweden (e-mails: \{daniel.gedon, antonio.horta.ribeiro, niklas.wahlstrom, thomas.schon\} @it.uu.se)}
}

\markboth{Submitted to 
IEEE Signal Processing Letters, No. XX, February, 2023 
}
{Shell \MakeLowercase{\textit{et al.}}: Bare Demo of IEEEtran.cls for IEEE Journals}
\maketitle

\begin{abstract}
Kernel principal component analysis (kPCA) is a widely studied method to construct a low-dimensional data representation after a nonlinear transformation. The prevailing method to reconstruct the original input signal from kPCA---an important task for denoising---requires us to solve a supervised learning problem. In this paper, we present an alternative method where the reconstruction follows naturally from the compression step. We first approximate the kernel with random Fourier features. Then, we exploit the fact that the nonlinear transformation is invertible in a certain subdomain. Hence, the name \textit{invertible kernel PCA (ikPCA)}. We experiment with different data modalities and show that ikPCA performs similarly to kPCA with supervised reconstruction on denoising tasks, making it a strong alternative. 
\end{abstract}


\IEEEpeerreviewmaketitle

\section{Introduction}

\IEEEPARstart{P}{rincipal} Component Analysis (PCA) involves finding a projection matrix $\mP$ that transforms a given input $\vx\in \R^\dimFeatures$ into a lower-dimensional representation $\vz = \mP\vx\in\R^\dimLatent$, with $\dimLatent<\dimFeatures$. Conversely, given a lower-dimensional representation, the original input space can be reconstructed with the inverse transformation $\hat{\vx} = \mP^\top \vz$. The data $\vx$ is often assumed to lie on a low-dimensional manifold. In such cases, PCA is beneficial since it enables the extraction of the most important features or directions of maximum variability in the data. The algorithm is optimal~\cite{shalev2014understanding} in the sense that there is no reconstruction matrix $\mU$ and reduction matrix $\mV$ such that the average distance between the original and reconstructed vector $\|\hat\vx - \mU \mV \vx\|_2$  is smaller than for $\mU = \mP^\top$ and $\mV=\mP$. 
Importantly, the matrix $\mP$ serves both as a tool for dimensionality reduction and as a means for  reconstructing the original input through its transpose $\mP^\top$.

Kernel PCA (kPCA) builds upon traditional PCA by enabling the study of the principal components after a nonlinear transformation~\cite{scholkopf1997kernel}. This allows for the generalization of the assumption that the data lies on a low-dimensional linear manifold to cases where this manifold is nonlinear. Traditional PCA might not be capable of retrieving useful low-dimensional representations $\vz$ in this scenario, but kPCA might succeed by using PCA after a nonlinear transformation $\Phi$ of the input $\vx$ into a (possibly infinite-dimensional) feature space $\gF$
\begin{equation}
    \label{eq:kpca_reduction}
    \vz = \mP \Phi(\vx).
\end{equation}
kPCA is indeed a natural and valuable idea. However, while the dimensionality reduction can be easily computed it is far from obvious how to obtain a reconstructed $\hat{\vx}$ from $\vz$.

This inverse reconstruction problem is known as the pre-image problem. Solutions are proposed based on gradient descent \cite{burges1996simplified}, nonlinear optimization \cite{mika1998kernel} or distance constraints in feature space \cite{kwok2003pre}.
%
The most widely disseminated solution by~\citet{bakir2004learning}, is to apply \eqref{eq:kpca_reduction} to construct a data set $\train=\{(\vx_i, \vz_i)\}_{i=1}^\numSamples$ consisting of original inputs and their low-dimensional representations. The goal is to find a nonlinear function $f$ that maps $\vz_i$ back to $\vx_i$. This approach is available, for instance, in scikit-learn~\cite{scikit_learn} or the multivariate statistics package for Julia~\cite{Julia_2017}. 
However, there are drawbacks to this approach: Unlike PCA, reconstruction is not an immediate by-product of kPCA and instead requires solving a supervised learning (SL) problem. Here, we denote this combination as kPCA+SL. Moreover, since the function~$f$ needs to be nonlinear, the supervised problem of finding the map between $\vz$ and $\vx$ usually results in a non-convex optimization problem. Indeed, direct nonlinear approaches---such as autoencoders \cite{bourlard1988auto,hinton1993autoencoders} and variational autoencoders \cite{kingma2013auto,rezende2014stochastic}---that concurrently implement dimensionality reduction and reconstruction, can yield significantly improved performance over kPCA+SL. While deep autoencoders are popular components of generative models, they require solving a non-convex optimization problem. Contrarily, kernel methods and PCA are well-understood and widely adopted preprocessing steps.

We propose a new formulation of kPCA that provides the reconstruction method as a direct by-product. The method works for any translational-invariant kernel. As we will discuss in Section~\ref{sec:background}, any such kernel can be approximated by a feature map of the type $\Phi(x) = \sigma(\mW \vx + \vb)$ with a nonlinearity $\sigma$ and $\mW \in \R^{\dimRFF \times \dimFeatures}$. Here, the dimensionality is reduced by the following sequence of computations
\begin{subequations}
    \label{eq:kpca}
    \begin{align}
        \valpha &= \mW \vx + \vb, \\
        \vbeta &= \sigma(\valpha),\\
        \vz &= \mP\vbeta.
    \end{align}
\end{subequations}

\begin{figure}
    \centering
    \includegraphics{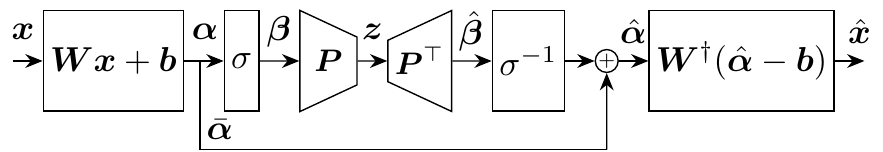}
    %
    \caption{Illustration of our invertible kernel PCA method.}
    \label{fig:method_ilustr}
\end{figure}

The method we propose involves inverting the operations step-by-step, as depicted in Fig.~\ref{fig:method_ilustr}. If a particular operation $\sigma(\valpha)$ cannot be inverted, we decompose the vector $\valpha$ into two components $\valpha-\bar\valpha$ and $\bar\valpha$, such that $\valpha-\bar\valpha$ belongs to a domain where $\sigma$ is invertible. We can use PCA to compress and decompress the first component, while the second component is bypassed. In this way, we avoid any nonlinear supervised problem and the reconstruction follows directly.


\section{Background}
\label{sec:background}

\noindent
Let, $k(\vx, \vy)$ be a positive semidefinite kernel
\begin{equation}
    \label{kernel_decomposition}
    k(\vx, \vy) = \langle\Phi(\vx), \Phi(\vy)\rangle = \sum_i \phi_i(\vx) \phi_i(\vy),
\end{equation}
where $\Phi(\vx) = (\phi_1(\vx), \phi_2(\vx), \dots)$ denotes a sequence of values  that maps the input into the feature space $\gF$.

\subsection{Kernel PCA}
For a set of observations $\{\vx_i\}_{i = 1}^\numSamples$ the empirical covariance matrix in $\gF\times \gF$ is given by
\begin{align*}
    \hat\mSigma = \frac{1}{\numSamples}\sum_{i = 1}^\numSamples \Phi(\vx_i) \Phi(\vx_i)^\top.
\end{align*}
The spectral decomposition of this matrix yields
\begin{align*}
    \hat\mSigma = \sum_i \lambda_i \vv_i \vv_i^\top,
\end{align*}
such that $\lambda_1 \ge \lambda_2 \ge \dots$. We define the projection into the first $\dimLatent$ components as
$\mP = \begin{bmatrix}
\vv_1 & \cdots & \vv_\dimLatent
\end{bmatrix}^\top$. For which we can obtain the lower dimensional representation ${\vz = \mP \Phi(\vx)}$.

\subsection{Infinite dimensional feature maps}

In practice, the kernel trick enables working with feature spaces of infinite dimension. The method we propose here is however intended for finite-dimensional feature spaces. Hence, when dealing with infinitely dimensional feature maps, we will resort to approximations. Specifically, we will use the feature map truncated to the first $\dimRFF$ components denoted as $\widetilde \Phi(\vx) = (\phi_1(\vx), \phi_2(\vx), \dots, \phi_\dimRFF(\vx))$ to approximate the kernel, meaning that we can write $k(\vx, \vy) \approx \langle\widetilde\Phi(\vx), \widetilde\Phi(\vy)\rangle$.

We follow the development of~\cite{rahimi_random_2008} using \textit{random Fourier features} to approximate a translation-invariant kernel, i.e. kernels of the form $k(\vx, \vy) = g(\vx-\vy)$. Bochner theorem guarantees that this kernel is continuous and positive semidefinite iff $g(\delta)$ is the Fourier transform of a probability distribution $p(\omega)$, possibly re-scaled. 

Take as an example the Gaussian kernel $k(\vx, \vy) = \exp{\frac{-\|\vx - \vy\|_2^2}{2}}$ which allows for the decomposition~\eqref{kernel_decomposition} only when considering an infinite dimensional feature space. To approximate this features space with random Fourier features, let $\mW \in \R^{\dimRFF \times \dimFeatures}$ be a matrix with random i.i.d. entries drawn from the distribution $p(\omega)$ and let $\vb\in \R^{\dimRFF}$ be a vector drawn i.i.d. from $\gU(-\pi, \pi)$. Then, we obtain
\begin{align*}
    \tilde \Phi(\vx) =\sqrt{2} \sin(\mW \vx + \vb),
\end{align*}
where $\sin$ is applied element-wise. It is proved in~\cite{rahimi_random_2008} that, $\langle \tilde \Phi(\vx), \tilde\Phi(\vy) \rangle$ converges uniformly to $k(\vx, \vy)$. Moreover, the convergence is exponentially fast in $\dimRFF$.

\section{Invertible kernel PCA}

\noindent
Let us consider feature maps of the type
\begin{align*}
    \Phi(\vx) = \sigma(\mW \vx + \vb),
\end{align*}
where $\mW \in \R^{\dimRFF \times \dimFeatures}$, $\vb \in \R^{\dimRFF}$ and $\sigma$ is a nonlinearity applied element-wise. The discussion in the previous section motivates how these feature maps can be used to approximate the space associated with any translational-invariant kernel. Next, we detail how to invert the operations, given that the dimensionality reduction was computed according to~\eqref{eq:kpca}. One of the key challenges is the fact that the activation function $\sigma$ is in general non-invertible. We describe our solution to deal with these problems next.

\subsection{Non-invertible activation functions}
In most cases of interest, the nonlinear function $\sigma: \R \rightarrow \R$, $\sigma: \alpha \mapsto \beta$ is not invertible in the entire domain $\R$, but it might be invertible in a subdomain $\mathcal{X} \subset \R$. Denote $\sigma_\mathcal{X}$ as the function $\sigma$ restricted to $\mathcal{X}$, then the inverse $\sigma_\gX^{-1}$ is well-defined. Let $\bar\alpha = \alpha - \sigma_\gX^{-1} \circ\sigma(\alpha)$.
Consider two examples: 
First, for the ReLU activation function $\beta = \sigma(\alpha) = \max(\alpha, 0)$, the invertible domain is $\mathcal{X} = [0, \infty)$. Thus, $\sigma_\gX^{-1}(\beta) = \beta$ and $\bar\alpha = \min(\alpha, 0)$. 
Second, for $\beta = \sigma(\alpha) = \sin \alpha$, as used in random Fourier features, we have that $\sigma$ is invertible in $\mathcal{X} = (-\pi/2, \pi/2]$. Thus, $\sigma_\gX^{-1}(\beta) = \arcsin \beta$ and $\bar\alpha =  (-1)^k  \alpha + \pi k$  for some $k\in\sZ$.


\subsection{ikPCA}

The reconstruction method inverts the operations in \eqref{eq:kpca} step-by-step. We can write 
\begin{subequations}
    \label{eq:reconstruction_ikpca_full}
    \begin{align}
        \widehat\vbeta  &= \mP^\top\vz,  \label{eq:reconstruction_ikpca}\\
        \widehat\valpha  &= \sigma_\gX^{-1}(\widehat\vbeta)  + \bar\valpha, \label{eq:inverting-nonlinear} \\
        \widehat{\vx} &= \text{arg}_{\vx}\min \|\mW \vx +\vb - \widehat\valpha \|_2^2 + \lambda \|\vx\|_2^2. \label{eq:reconstruction_ridge}
    \end{align}
\end{subequations}
The first step \eqref{eq:reconstruction_ikpca} inverts the dimensionality reduction, and is motivated by the same reasoning as PCA: 
the projection matrix ${\mP}$ is such that the reconstruction error $\|\hat\vbeta - \mP^\top \vz\|_2$ is minimal. The second step \eqref{eq:inverting-nonlinear}, inverts the nonlinear function $\sigma$ on the subdomain $\gX$ and adds the bypassed non-invertible part $\bar\valpha$. Finally, the last step \eqref{eq:reconstruction_ridge} inverts the linear map $\vx \mapsto \mW \vx  +\vb$ by solving a Ridge regression problem. Notice that for $\lambda \rightarrow 0^{+}$ the last step reduces to $\widehat{\vx} = \mW^{\dagger} (\widehat\valpha - \vb)$, where $\mW^{\dagger}$ is the pseudo-inverse of $\mW$, as in Fig.~\ref{fig:method_ilustr}.


\section{Numerical examples}

In this section, we outline the experiments to evaluate the performance of the proposed ikPCA method. We focus on the task of denoising inputs of various modalities. Quantitatively, we evaluate the mean square error (MSE) between the de-noised test signal and the true non-noisy test signal. We compare ikPCA with PCA and kPCA+SL due to their structural similarity. To ensure a fair comparison, we did not consider denoising autoencoders, which present hierarchical, deep models. In the discussion, we detail how our methodology could be extended to neural networks and a setup that could be better compared with autoencoders.

In all experiments, we consider the Gaussian kernel (or its random Fourier feature approximation) and present the results in terms of mean and standard deviation over 20 random runs. Parameters of the methods which are fixed in an experiment were optimized through hyperparameter grid search. For reproducibility, we release our code publicly\footnote{Code is available at \url{https://github.com/dgedon/invertible_kernel_PCA}}.

\subsection{Synthetic toy data: s-curve}
\begin{figure}
    \centering
    \includegraphics{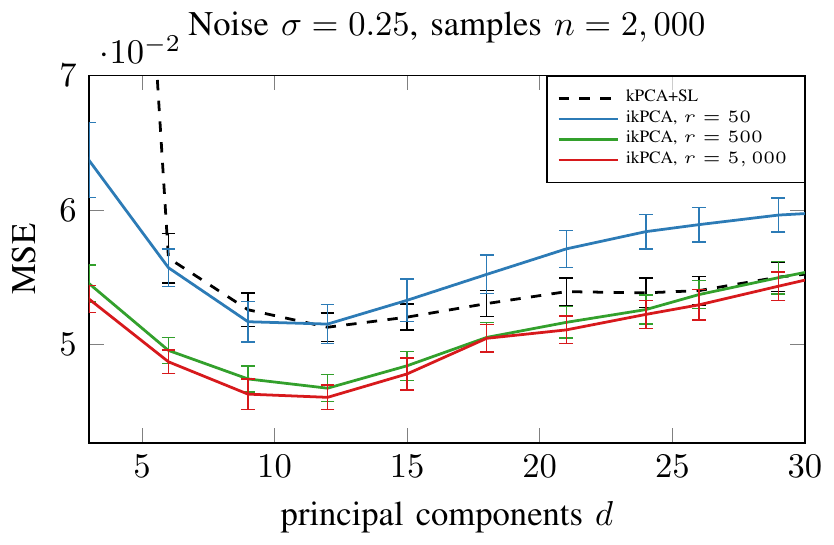}
    %
    \caption{S-curve toy example. Reconstruction MSE for a different number of random features chosen for ikPCA.}
    \label{fig:scurve rff 025}
\end{figure}

We generated synthetic 3-dimensional data points in the shape of the letter `S' using the s-curve toy problem. For training and testing, we generate $\numSamples=2,000$ data points and add Gaussian noise with $\sigma=0.25$. For kPCA+SL we set the kernel width $\gamma=1$ and the reconstruction Ridge strength $\lambda=1$; for ikPCA we set $\gamma=0.5$ and $\lambda=1$. The results are presented in Fig.~\ref{fig:scurve rff 025}. Our proposed ikPCA method is capable of denoising the data in a comparable manner to kPCA+SL. For this problem, we observe that as few as $\dimRFF=50$ random Fourier features were sufficient to match the performance of kPCA+SL. This provides ikPCA with a computational advantage over kPCA+SL, which needs to invert a $\numSamples\times\numSamples$ matrix.

\subsection{USPS Images}
\begin{figure}
    \centering
    \includegraphics{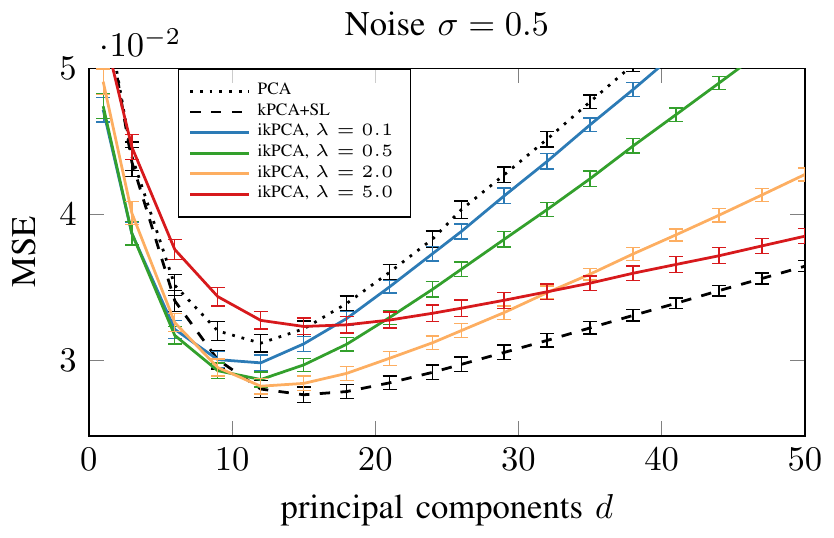}
    %
    \caption{USPS data. Effect of regularization parameter $\lambda$.}
    \label{fig:usps regularize 05}
\end{figure}

We utilize the USPS data set which contains handwritten digits in a greyscale format of size $16\times16$ and add Gaussian noise with $\sigma=0.5$. We use $\numSamples=1,000$ images for training and $400$ images for testing. For kPCA+SL we set $\gamma=5\cdot10^{-3}$ and $\lambda=10^{-2}$; for ikPCA we use $30,000$ random Fourier features and set $\gamma=10^{-4}$. In Fig.~\ref{fig:usps regularize 05}, we vary the regularization parameter $\lambda$ of ikPCA. Again, our findings show that ikPCA performs similarly to kPCA+SL for an optimal number of principal components $\dimLatent$. The figure further suggests that Ridgeless reconstruction behaves comparably in performance to PCA. However, excessive regularization negatively affects the overall reconstruction performance.

\begin{figure}
    \centering
    \includegraphics{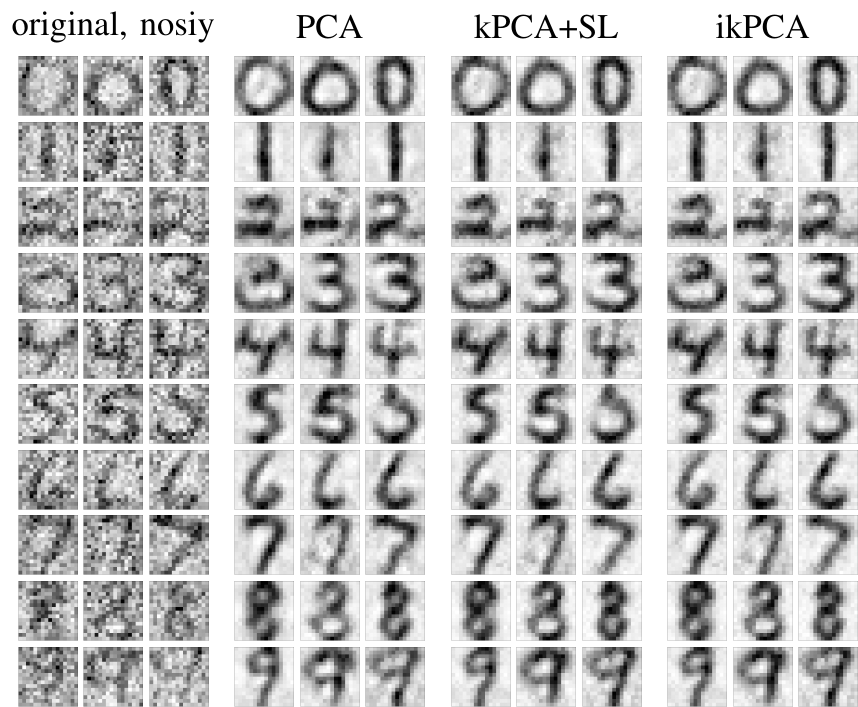}
    %
    \caption{USPS reconstruction. }
    \label{fig:usps reconstruction 025}
\end{figure}

Fig.~\ref{fig:usps reconstruction 025} displays image denoising results for noise scale $\sigma=0.25$. For ikPCA we chose the optimal regularization $\lambda=1.3$. The number of principal components $\dimLatent$ is chosen for each method such that the MSE is minimised. All methods demonstrate visually comparable image denoising capabilities, which is supported by the difference in MSE from Fig.~\ref{fig:usps regularize 05}.

\subsection{Electrocardiogram}
The electrocardiogram (ECG) is a routine, medical test that records the heart's electrical activity, typically used to diagnose various heart conditions. However, noise measured during the recording can complicate the diagnosis. Several methods have been proposed to de-noise the ECG signal. For comparisons, two approaches have been suggested: (1) artificially adding noise to the signal and comparing with the original one itself \cite{sameni2007nonlinear,xiong2016ecg,chiang2019noise}, or (2) de-noise the existing signal and comparing it to the mean beat as the noise-free reference \cite{castells2007principal,johnstone2009sparse}. We choose the latter approach to account for real-world noise scenarios.

We utilize ECGs from the China Physiological Signal Challenge 2018 (CPSC)\footnote{Data is available at \url{http://2018.icbeb.org/Challenge.html}} which contains data between 6 and 60 seconds long \cite{liuCPSC2018}. From the 918 ECGs with no abnormalities, we selected the longest recordings and focused on a single lead in this example. To extract the beats, we first remove baseline wander with a high-pass filter. Then, we identify the R-peaks \cite{wfdb_2023}, resample the interval between each peak to 512 samples and finally locate the R-peak at the 150th sample following the preprocessing approach of \citet{johnstone2009sparse}.

\begin{figure}
    \centering
    \includegraphics{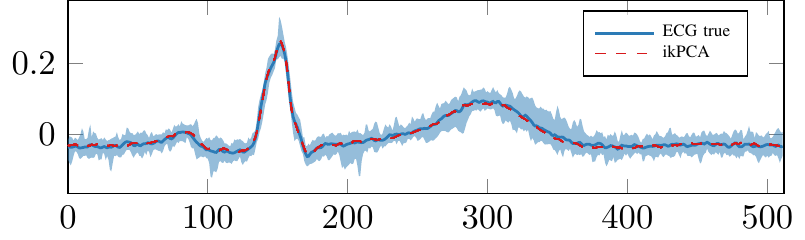}
    %
    \caption{Denoising of ECG beats from lead I. The blue area marks the min/max values of the 21 test beats. The red dashed lines show all test reconstructions with ikPCA.}
    \label{fig:ECG reconstruction}
\end{figure}

We extracted 70 beats from the selected ECG; 49 for training and 21 for testing. Applying kPCA+SL with $\gamma=10$ and $\lambda=15$, and ikPCA with $\gamma=5\cdot10^{-5}$ and $\lambda=10$, along with the minimum of 512 random Fourier features $\dimRFF$, we achieved perfect signal denoising using only the first component, as shown in Fig.~\ref{fig:ECG reconstruction}. Quantitatively over 500 simulations, the MSE for ikPCA was $2.6\pm0.8\cdot10^{-5}$, similar to that of kPCA+SL, while PCA had a slightly higher MSE.

\section{Computational considerations}
The computational complexity is not increased by adding the reconstruction stage for ikPCA. The reason is that the cost of obtaining the reconstruction is smaller than the cost of the kPCA decomposition (whenever the input dimension $\dimFeatures$ is lower than the number of samples $\numSamples$). However, our method requires the kernel map to be approximated by~$\dimRFF$ random Fourier features. 
When $\dimRFF < \numSamples$ this might reduce the computational cost, but when $\dimRFF > \numSamples$ the computational cost is increased by a factor of $\dimRFF/\numSamples + 1$ compared to that of kPCA.

\paragraph{kPCA computational cost} 
Some kernels have closed forms that can be computed in $\bigO(1)$ operations. The cost for kPCA is then dominated by the inversion of the Gram matrix $\mK$ which requires $\bigO(\numSamples^3)$ operations. The Gram matrix being the matrix with entry $(i, j)$ equal to $k(\vx_i, \vx_j)$.

\paragraph{Computation cost of PCA in the feature space}
In ikPCA we approximate the kernels with finite, $\dimRFF$-dimensional features $\Phi(\vx) = (\phi_1(\vx), \dots, \phi_\dimRFF(\vx))$, and perform PCA on the covariance matrix of the features $\hat\mSigma = \frac{1}{\numSamples}\sum_i^\numSamples\Phi(\vx_i) \Phi(\vx_i)^\top$. Computing the entries of the matrix and its spectral decomposition requires $\bigO(\dimRFF^2 \numSamples + \dimRFF^3)$ operations. Hence, for $\dimRFF < \numSamples$, approximating the kernel and computing the spectral decomposition of $\hat\mSigma$ might be computationally more efficient than working directly with the Gram Matrix $\mK$ as in kPCA.

However, if $\dimRFF > \numSamples$ this advantage is diminished and it can be efficient to work with the Gram matrix $\mK= \Phi (\mX)^\top \Phi (\mX)$ instead of $\hat\mSigma$. $\mK$ has the same (nonzero) eigenvalues as $n\hat\mSigma$, and its eigenvectors multiplied by $\Phi (\mX)$ yield the eigenvectors of $n\hat\mSigma$. The cost in this formulation is $\bigO(\dimRFF \numSamples^2 + \numSamples^3)$. Therefore, the cost is a factor of $\dimRFF/\numSamples + 1$ times higher than the cost obtained for kernels with a closed-form solution.

According to Claim~1 in \cite{rahimi_random_2008}, $\dimRFF = \Omega\left( \frac{\dimFeatures}{\epsilon^2} \log \frac{D}{\epsilon}\right)$ random Fourier features are required to ensure an approximation error smaller than $\epsilon$ on a space of diameter $D$. Thus, $\dimRFF$ grows linearly with the input dimension $\dimFeatures$. In the case of the s-curve example, $\dimRFF \ll \numSamples$, whereas in the USPS example, $\dimRFF > \numSamples$, due to large $\dimFeatures$ and our method's computational advantage is lost. For high-dimensional data like the latter, Nyström approximations \cite{williams2000using} could be used and be more efficient in terms of $\dimRFF$ \cite{yang2012nystrom}.

\paragraph{Cost of reconstruction} The reconstruction cost in ikPCA is dominated by the cost of solving the optimization problem~\eqref{eq:reconstruction_ridge}. For this, we require computing the SVD of $\mW$ one single time with a cost of $\bigO(\dimFeatures^3 + \dimFeatures^2 \dimRFF)$. The cost of solving the reconstruction is then $\bigO(\dimFeatures \dimRFF + \dimLatent \dimRFF)$ for each new $\hat\valpha$.

\section{Conclusion and discussion}

We propose an invertible version of kPCA+SL. While the traditional approach solves a supervised problem to map back from the latent space to the input space, our method obtains this mapping naturally. We approximate the kernel transformation with random Fourier features $\Phi(x)=\sigma(\mW x + \vb)$. Although the nonlinear function $\sigma$ might not be invertible, we observe that it can be inverted in a subdomain. We can exploit this observation by decomposing its input into invertible and non-invertible parts and bypassing the second. We show the effectiveness of our approach for denoising in three examples: an s-curve toy problem, the USPS image data set and ECGs.

We compare our method with symmetric kPCA+SL. Symmetry implies here that the kernel for compression and reconstruction are defined identically, which is motivated by implementations in common frameworks \cite{scikit_learn,Julia_2017}. However, the method in \cite{bakir2004learning} is not limited to this by design. Conversely, ikPCA is required to have a symmetric setup due to the natural inversion of the nonlinear transformation in the reconstruction. While our method aligns well with kPCA+SL in the numerical experiments we presented, it remains uncertain how it would compare against a well-tuned non-symmetric kPCA+SL.




Despite the simplicity of our method, there is a wide array of possible extensions. To extend the representational power, we can stack multiple layers of $\Phi(x)=\sigma(\mW x + \vb)$ transformations in a hierarchical way. Hence, we obtain a structure which is closer to that of a deep autoencoder. This may allow drawing further connections between the theoretically well-established kernel regime and neural networks. 
In a similar direction, we can view the random Fourier features in our method as an untrained, single-layer neural network. Extending our method to trained neural networks would allow performing reconstruction tasks naturally without re-training.
Finally, we experiment with underparameterized data (USPS example with $\dimFeatures/\numSamples \approx 0.25$) and overparameterized data (ECG example with $\dimFeatures/\numSamples \approx 10$). This fact, combined with the use of a high number of random Fourier features, raises questions about overparameterization and benign overfitting of denoising models \cite{radhakrishnan2018memorization,Bartlett2020}. 



\section*{Acknowledgment}
The computations were enabled by the supercomputing resource Berzelius provided by National Supercomputer Centre at Linköping University and the Knut and Alice Wallenberg foundation.

\printbibliography

\ifarxiv
\clearpage
\pagenumbering{roman}
\setcounter{page}{1} 
\beginsupplement
\onecolumn
\appendix

\subsection{Additional results on s-curve data set}
\label{app:s-curve}
The s-curve is generated by the following set of equations where the variable $t$ is often used as a label\footnote{See also \url{https://scikit-learn.org/stable/modules/generated/sklearn.datasets.make_s_curve.html}}. For our purposes, we do not require labels but are only concerned with inputs $\vx$. Additive Gaussian noise $\vv\sim\gN\left(0,\sigma \mI_3\right)$ is added to $\vx$.
\begin{align*}
    t &\sim \gU\left(-\frac{3}{2}\pi,\frac{3}{2}\pi\right), \\
    x_1 &= \sin{t}, \\
    x_2 &\sim \gU\left(0,2\right), \\
    x_3 &= \mathrm{sign}(t) \left(\cos{t}-1\right).
\end{align*}
Fig.~\ref{fig-app:scurve visualization} shows a visualisation of the s-curve data set. Fig~\ref{fig-app:scurve rff} is an extension of Fig~\ref{fig:scurve rff 025} for a larger set of random Fourier features $\dimRFF$ and for a second level of additive noise. For this data set, fewer random Fourier features are necessary than training data points. Hence, our method is numerically faster. Already $\dimRFF=500$ features (less than $1/4$ of the number of samples $\numSamples=2,000$) are sufficient for the optimal performance curve. We observe that for larger noise values, ikPCA even outperforms kPCA+SL and PCA and that the effect of $\dimRFF$ is less pronounced.

\begin{figure}[H]
    \centering
    \includegraphics{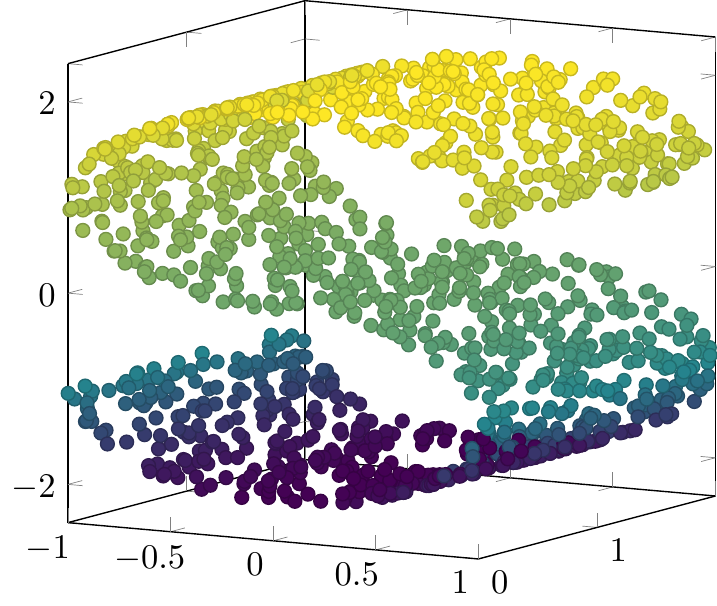}
    %
    \caption{Visualization of the s-curve data set. The colour indicates the regression label $t$.}
    \label{fig-app:scurve visualization}
\end{figure}

\begin{figure}[H]
    \centering
    \begin{subfigure}[b]{0.48\textwidth}
        \centering
         \includegraphics{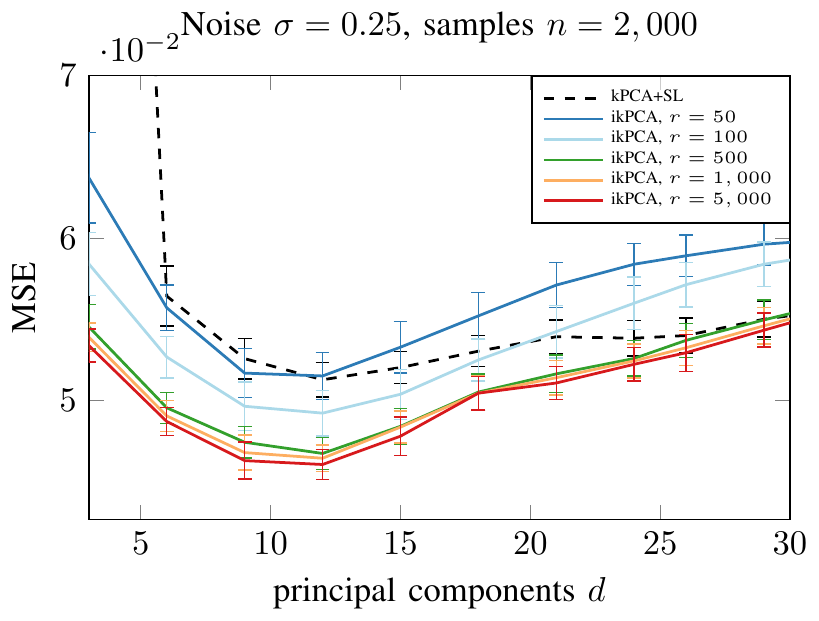}
        %
     \end{subfigure}
     \hfill
     \begin{subfigure}[b]{0.48\textwidth}
        \centering
        \includegraphics{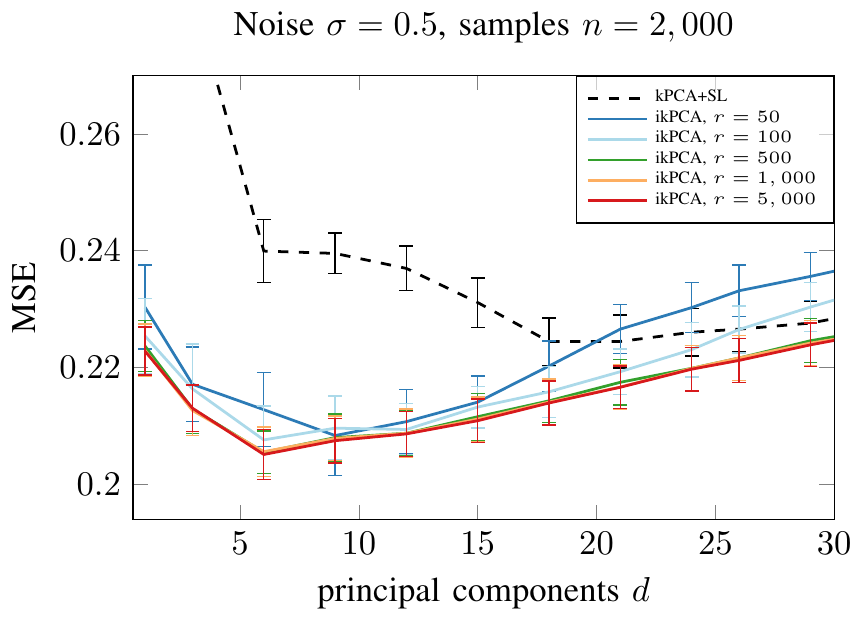}
        %
    \end{subfigure}
    \caption{Effect of the number of random features components on reconstruction MSE. Fig.~\ref{fig:scurve rff 025} is a modified version of the left figure here.}
    \label{fig-app:scurve rff}
\end{figure}

\newpage
\subsection{Additional results on USPS data set}
\label{app:usps}
For the following plots, the hyperparameters for kPCA+SL (i.e. kernel width $\gamma$ and regularization strength $\lambda$) were selected such that the lowest reconstruction MSE was achieved. A grid search was utilized. For all results mean (and in error plots also standard deviation) over 20 random runs are presented.

Fig.~\ref{fig-app:USPS rff} explores the effect of the number of random Fourier features $\dimRFF$ for this data set. We observe that generally more random Fourier features yield asymptotically better results. Furthermore, we note that our method ikPCA approaches kPCA+SL for $\dimRFF\rightarrow\infty$ as suggested by the approximation of the kernel.

Fig.~\ref{fig-app:USPS regularization} explores the effect of the regularization parameter $\lambda$ for the reconstruction in our ikPCA method. We observe that an optimal trade-off has to be found. For $\lambda\rightarrow0^+$, ikPCA approaches the performance of PCA. Conversely, for large values of $\lambda$, the problem becomes over-regularized and does not generalize anymore.

Fig.~\ref{fig-app:USPS noise 3d} shows the combined effect of the additive noise level and the number of principal components $\dimLatent$ chosen for the latent space. The number of components with the lowest MSE for each method is shown in the left plot of Fig.~\ref{fig-app:USPS noise 2}. We observe that a larger noise value leads to a lower number of optimal principal components $\dimLatent$, which is justified as the noise level dominates a larger portion of singular values. Fig.~\ref{fig-app:USPS noise 2} subsequently shows the MSE values of all three methods when choosing the optimal number of principal components $\dimLatent$. We observe that the MSE for optimal tuned methods in this data set is similar for all methods and noise levels.

Fig.~\ref{fig-app:USPS reconstruction} is a reconstruction of USPS images for two different noise levels when choosing optimal hyperparameters for all methods. As the quantitative comparison in the right plot of Fig.~\ref{fig-app:USPS noise 2} suggests, the reconstructions are also qualitatively similar.

\begin{figure}[H]
     \centering
     \begin{subfigure}[b]{0.45\textwidth}
         \centering
         \includegraphics{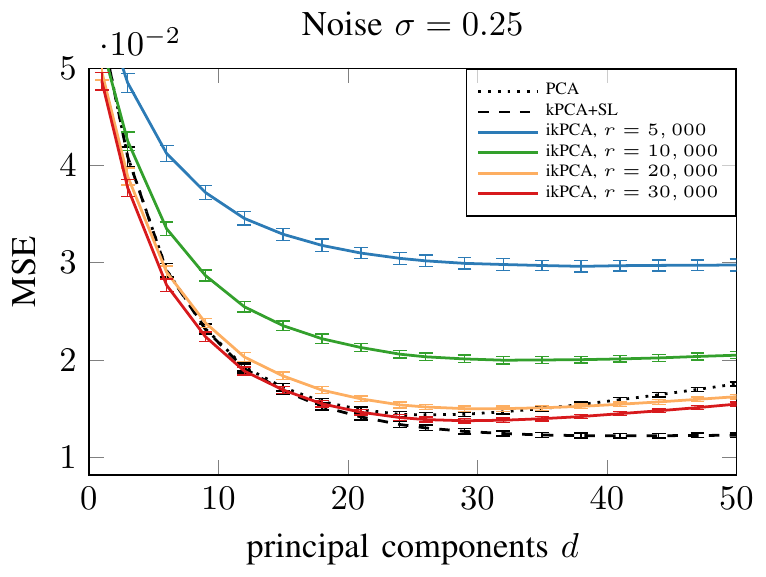}
         %
     \end{subfigure}
     \hfill
     \begin{subfigure}[b]{0.45\textwidth}
         \centering
         \includegraphics{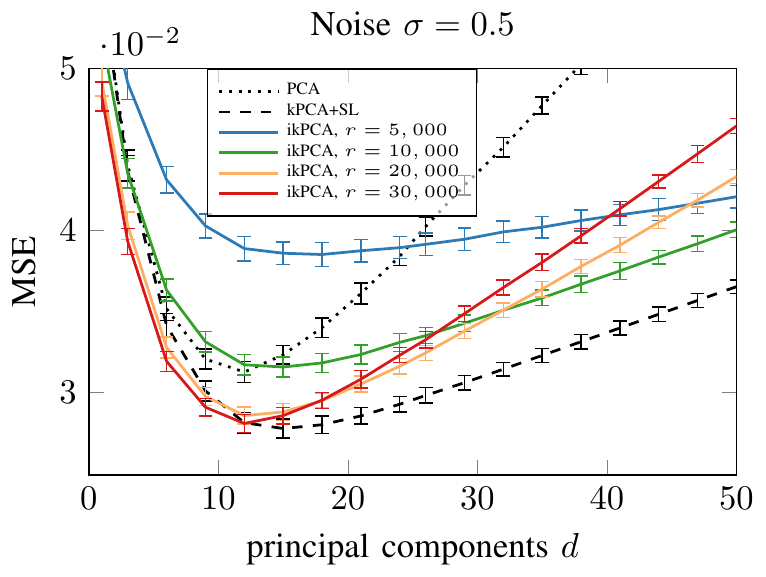}
         %
     \end{subfigure}
    \caption{Effect of different number of random feature components on reconstruction MSE. }
    \label{fig-app:USPS rff}
\end{figure}

\begin{figure}[H]
     \centering
     \begin{subfigure}[b]{0.45\textwidth}
         \centering
		 \includegraphics{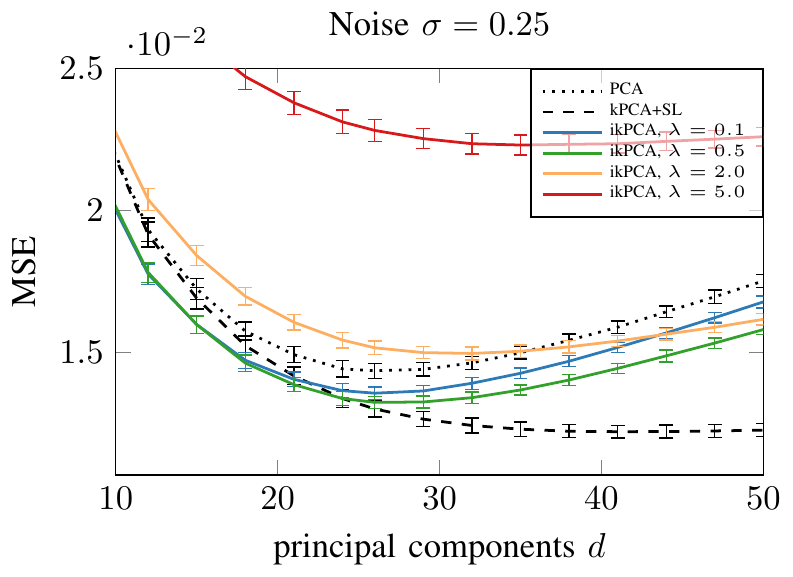}
         %
     \end{subfigure}
     \hfill
     \begin{subfigure}[b]{0.45\textwidth}
         \centering
         \includegraphics{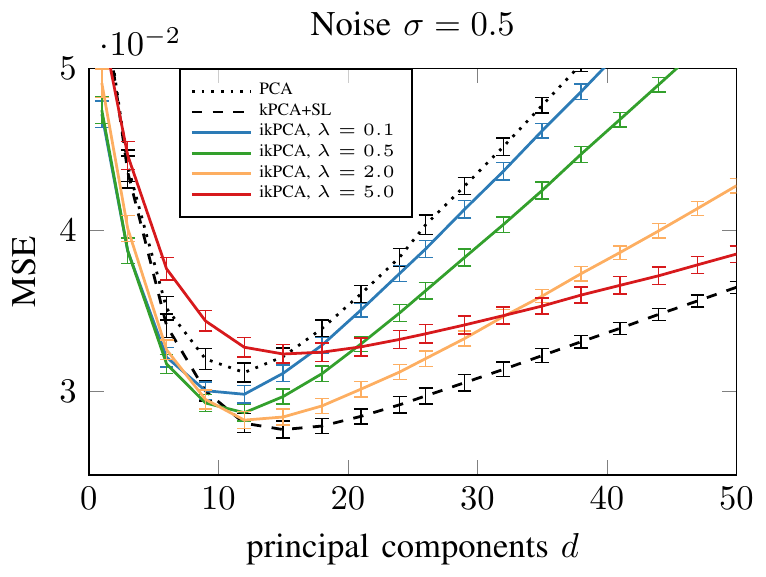}
         %
     \end{subfigure}
    \caption{Effect of regularization parameter $\lambda$ on reconstruction MSE. An optimal value has to be chosen. Right plot is a repetition of Fig.~\ref{fig:usps regularize 05}.}
    \label{fig-app:USPS regularization}
\end{figure}

\begin{figure}[H]
    \centering

    \begin{subfigure}[b]{0.3\textwidth}
   		\includegraphics{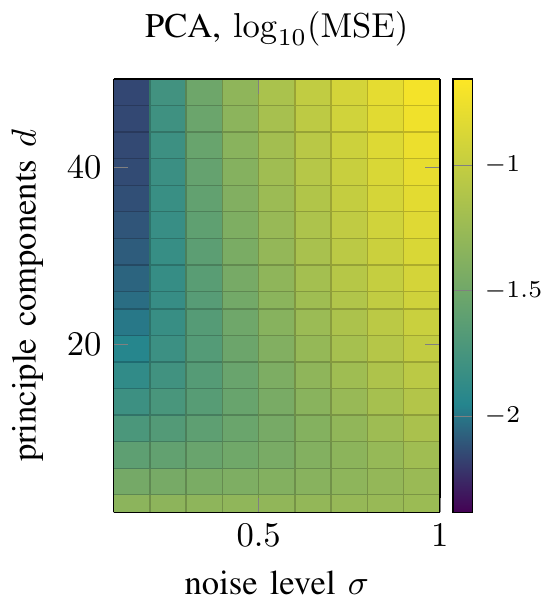}
    \end{subfigure}
	\hfill
	\begin{subfigure}[b]{0.3\textwidth}
		\includegraphics{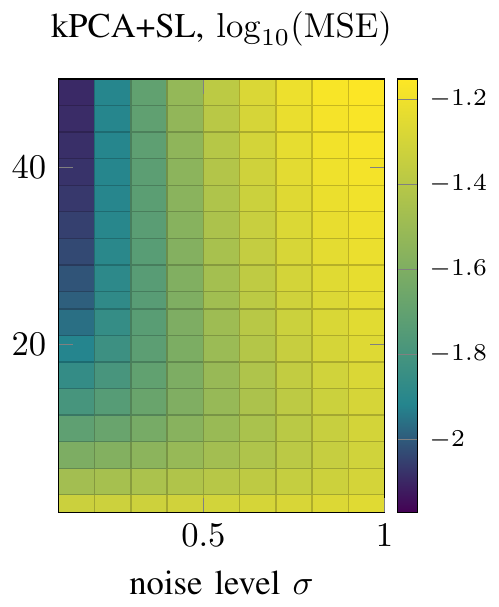}
	\end{subfigure}
	\hfill
	\begin{subfigure}[b]{0.3\textwidth}
		\includegraphics{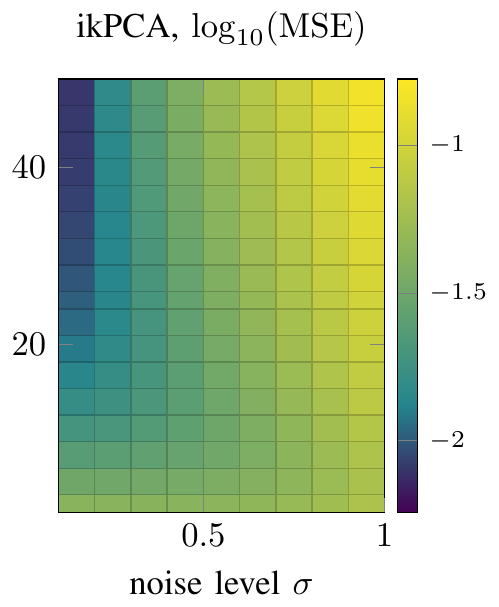}
	\end{subfigure}
	\hfill
    \caption{Combined effect of noise and latent space dimension on reconstruction MSE.}
    \label{fig-app:USPS noise 3d}
\end{figure}

\begin{figure}[H]
     \centering
     \begin{subfigure}[b]{0.45\textwidth}
         \centering
         \includegraphics{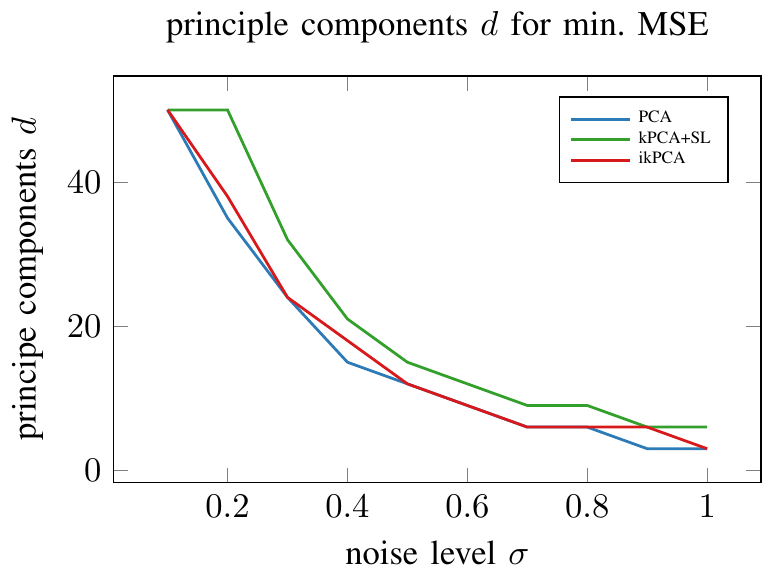}
         %
     \end{subfigure}
     \hfill
     \begin{subfigure}[b]{0.45\textwidth}
         \centering
         \includegraphics{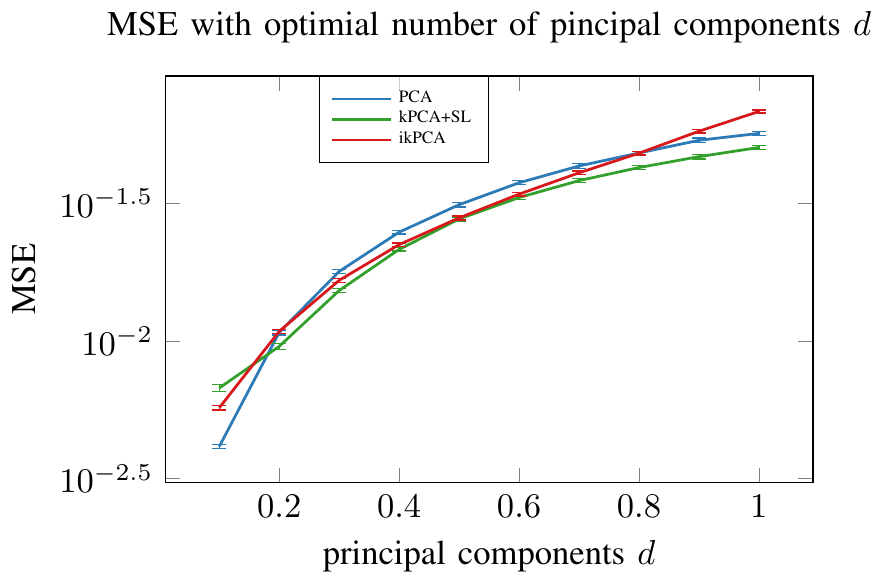}
         %
     \end{subfigure}
    \caption{Effect of noise. (Left) The best number of components to achieve the lowest MSE for a certain noise level. (Right) MSE of the three methods for reconstruction choosing the optimal number of components.}
    \label{fig-app:USPS noise 2}
\end{figure}

\begin{figure}[H]
     \centering
     \begin{subfigure}[b]{0.45\textwidth}
         \centering
         \includegraphics{figuresTikz/usps_reconstruction_025}
         %
         \caption{Noise level $\sigma=0.25$}
     \end{subfigure}
     \hfill
     \begin{subfigure}[b]{0.45\textwidth}
         \centering
         \includegraphics{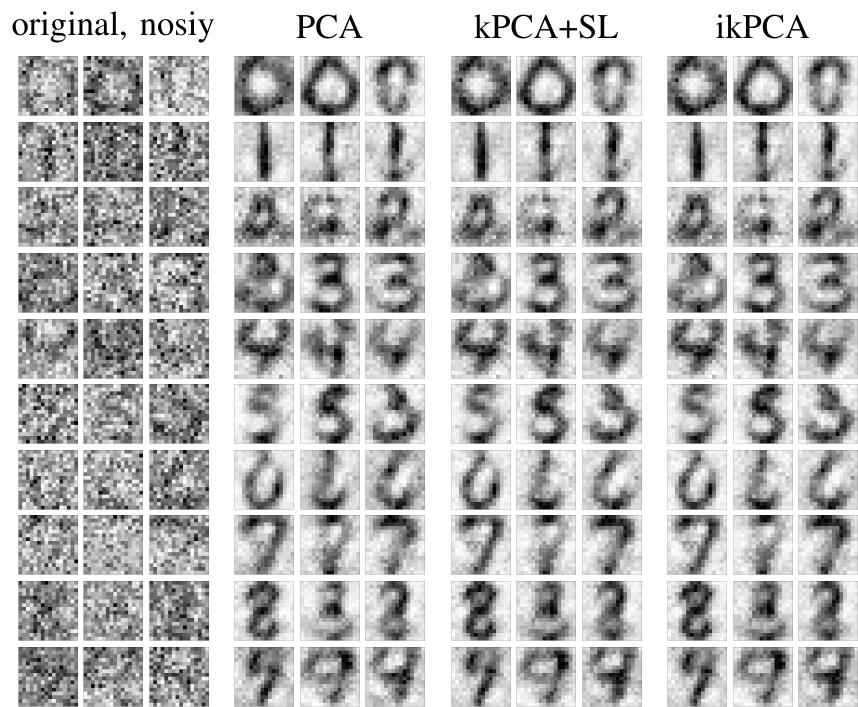}
         %
         \caption{Noise level $\sigma=0.5$}
     \end{subfigure}
    \caption{Reconstruction with different methods. Optimal hyperparameters were chosen for each method to achieve the lowest MSE. Left plot is a repetition of Fig.~\ref{fig:usps reconstruction 025}.}
    \label{fig-app:USPS reconstruction}
\end{figure}

\newpage
\subsection{Additional results on ECG data}
\label{app:ecg}

Fig.~\ref{fig-app:ECG reconstruction} shows two more examples of reconstructing ECG signals, complementing Fig.~\ref{fig:ECG reconstruction}. The same hyperparameters as in the main text are chosen. In the right plot, we can see that for PCA some reconstructions (red dashed lines) are not optimal, i.e. close to the ground truth line. This leads to a significantly higher MSE. Both kPCA+SL and ikPCA perform similarly.

Tab.~\ref{tab:ecg mse} compares the MSE values over 500~simulations with different train/test splits for the three ECG traces in Fig.~\ref{fig-app:ECG reconstruction}. We observe that ikPCA and kPCA+SL perform similarly in terms of MSE, while PCA has a slightly higher MSE. Hence, ECG denoising is not as good with a purely linear model.

\begin{figure}[H]
     \centering
     \begin{subfigure}[b]{0.48\textwidth}
         \centering
         \includegraphics{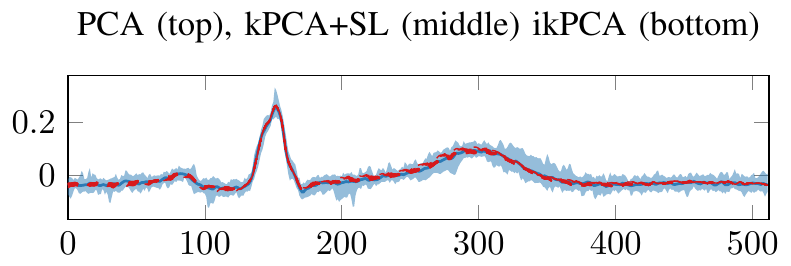}
         \includegraphics{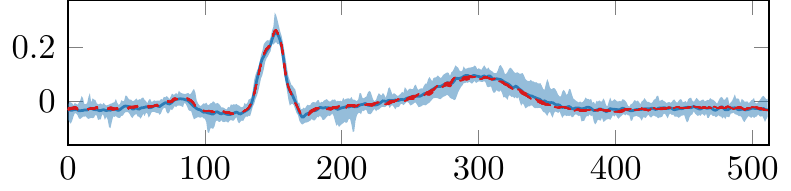}
         \includegraphics{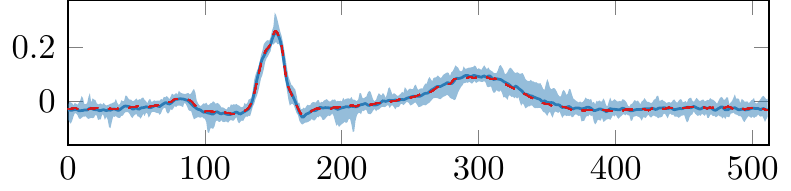}
         %
         \caption{Lead I of an ECG consisting of 70 beats (49/21 for training/test).}
         \label{subfigure a}
     \end{subfigure}
     \hfill
     \begin{subfigure}[b]{0.48\textwidth}
         \centering
         \includegraphics{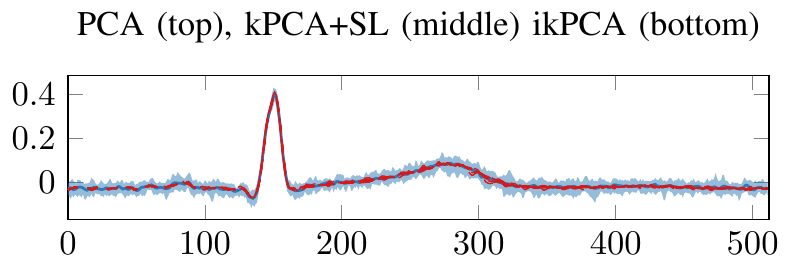}
         \includegraphics{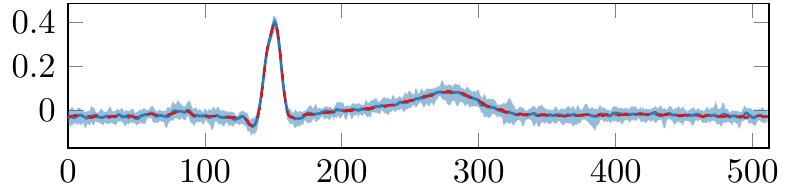}
         \includegraphics{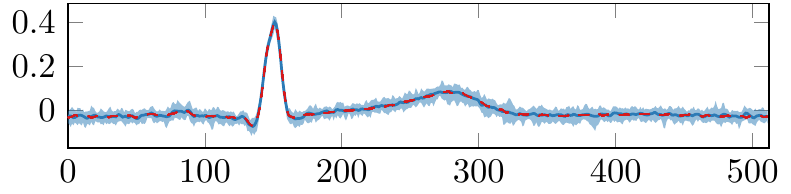}
         %
		 \caption{Lead I of an ECG consisting of 62 beats (43/19 for training/test).}
     \end{subfigure}
	 \vskip\baselineskip
	 \begin{subfigure}[b]{0.48\textwidth}
	 	\centering
        \includegraphics{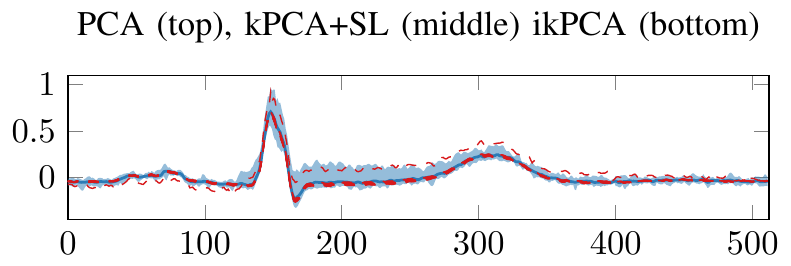}
        \includegraphics{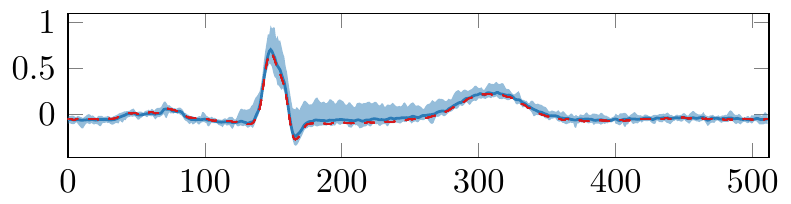}
        \includegraphics{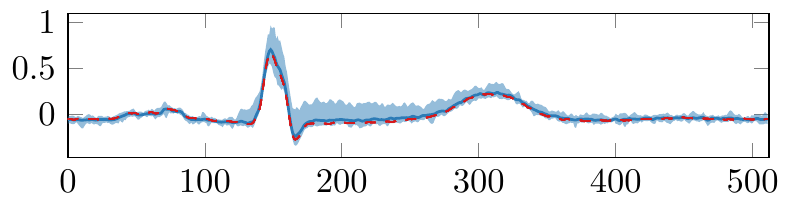}
        \caption{Lead II of an ECG consisting of 40 beats (28/12 for training/test).}
	 \end{subfigure}
	 \hfill
	 \begin{subfigure}[b]{0.48\textwidth}
	 	\centering
	 \end{subfigure}
    \caption{More ECG reconstruction results. Fig.~\ref{subfigure a} is the same example as in Fig.~\ref{fig:ECG reconstruction}.}
    \label{fig-app:ECG reconstruction}
\end{figure}

\begin{table}[H]
	\centering
	\begin{tabular}{c|ccc|c}
		& PCA & kPCA+SL & ikPCA & [unit] \\
	\toprule
	ECG (a)	& $4.00\pm1.47$ & $2.78\pm0.74$ & $2.57\pm0.79$ & [$10^{-5}]$  \\
	ECG (b)	& $3.20\pm0.35$ & $2.38\pm0.29$ & $2.32\pm0.31$ & [$10^{-5}]$  \\
	ECG (c) & $8.37\pm5.93$ & $2.43\pm1.63$ & $2.27\pm1.49$ & [$10^{-4}]$  \\
	\end{tabular}
	\caption{Reconstruction MSE for different ECGs.}
	\label{tab:ecg mse}
\end{table}
\fi

\end{document}